\documentclass[letterpaper, 10pt, conference]{ieeeconf}
\usepackage{cite}
\usepackage{amsmath,amssymb,amsfonts}
\usepackage{algorithmic}
\usepackage{graphicx}
\usepackage{textcomp} 
\usepackage{xcolor}
\usepackage{csquotes}
\usepackage{makecell}
\usepackage{color}
\usepackage{multirow}%
\usepackage{comment}
\usepackage{hyperref}
\newtheorem{definition}{\textbf{Definition}}
\newtheorem{theorem}{\rm\textbf{Theorem}}
\IEEEoverridecommandlockouts
\newtheorem{remark}{Remark}
\DeclareUnicodeCharacter{202F}{\,}

\def\BibTeX{{\rm B\kern-.05em{\sc i\kern-.025em b}\kern-.08em
    T\kern-.1667em\lower.7ex\hbox{E}\kern-.125emX}}

\begin{document}
\title{{\LARGE \textbf{Optimal Control of Connected Automated Vehicles with Event-Triggered Control Barrier Functions: a Test Bed for Safe Optimal Merging}}}
\author{Ehsan Sabouni$^{*,1}$, H.M. Sabbir Ahmad$^{*,1}$, Wei Xiao$^2$, Christos G. Cassandras$^1$ and Wenchao Li$^1$
\thanks{$^*$These authors contributed equally to this paper.}
\thanks{$^1$Division of Systems Engineering and Department of Electrical \& Computer Engineering, Boston University, Boston, MA, USA, \texttt{{\small \{esabouni, sabbir92, cgc, wenchao\}@bu.edu}}}
\thanks{$^2$Computer Science \& Artificial Intelligence Laboratory, Massachusetts Institute of Technology, Cambridge, MA, USA \texttt{{\small \{weixy@mit.edu\} }}}
\thanks{This work was supported in part by NSF under grants ECCS-1931600,
DMS-1664644, CNS-1645681, CNS-2149511, by AFOSR under grant FA9550-19-1-0158,
by ARPA-E under grant DE-AR0001282, by the MathWorks and by NPRP grant
(12S-0228-190177) from the Qatar National Research Fund, a member of
the Qatar Foundation (the statements made herein are solely the responsibility
of the authors).}
}%

\maketitle

\begin{abstract}                
We address the problem of controlling Connected and Automated Vehicles (CAVs) in conflict areas of a traffic network subject to hard safety constraints. It has been shown that such problems can be solved through a combination of tractable optimal control problems and Control Barrier Functions (CBFs) that guarantee the satisfaction of all constraints. These solutions can be reduced to a sequence of Quadratic Programs (QPs) which are efficiently solved on line over discrete time steps. However, guaranteeing the feasibility of the CBF-based QP method within each discretized time interval requires the careful selection of time steps which need to be sufficiently small. This creates computational requirements and communication rates between agents which may hinder the controller's application to real CAVs. In this paper, we overcome this limitation by adopting an event-triggered approach for CAVs in a conflict area such that the next QP is triggered by properly defined events with a safety guarantee. We present a laboratory-scale test bed we have developed to emulate merging roadways using mobile robots as CAVs which can be used to demonstrate how the event-triggered scheme is computationally efficient and can handle measurement uncertainties and noise compared to time-driven control while guaranteeing safety.
\end{abstract}

\section{Introduction}



The effective traffic management of Connected and Automated Vehicles (CAVs) through control and coordination has brought the promise of resolving long-lasting problems in transportation networks such as accidents, congestion, and unsustainable energy consumption \cite{deWaard09},\cite{Schrank20152015UM},\cite{kavalchuk2020performance}. To date, both centralized \cite{Liu_01, Xiao_05} and decentralized \cite{Xu_02, Chen_01, Han_01} methods have been proposed to tackle the control and coordination problem of CAVs in conflict area such as intersections, roundabouts, and merging roadways; an overview of such methods may be found in \cite{7562449}. In decentralized methods, as opposed to centralized ones, each CAV is responsible for its own on-board computation with information from other vehicles limited to a set of neighbors \cite{7313484}. One approach is to formulate a constrained optimal control problem 
jointly minimizing travel time through conflict areas and energy consumption, e.g., for optimal merging \cite{Hussain18} or crossing
 intersections \cite{2020LiIntersection}.
The complexity of obtaining the solution, however, necessitates the use of online control methods like Control Barrier Functions (CBFs) \cite{CBF_QP(2017)}, \cite{xiao_01} and Model Predictive Control (MPC) techniques \cite{Rawlings2018} for real-world applications.

An approach combining optimal control solutions with CBFs (termed OCBF) was recently presented in \cite{Xiao_03}. In this combined approach, the solution of a \emph{tractable} optimal control problem is used as a reference control. Then, the resulting control reference trajectory is optimally tracked subject to
a set of CBF constraints that ensure the satisfaction of all constraints of the original optimal control problem. Finally, this optimal tracking problem is efficiently solved by discretizing time and solving a simple Quadratic Problem (QP) at each discrete time step over which the control input is held constant \cite{CBF_QP(2017)}. 
However, the control update interval in the time discretization process must be sufficiently small in order to always guarantee that every QP is feasible. In practice, such feasibility can be often seen to be violated. 

One way to remedy this problem is to use an \emph{event-triggered} scheme instead. A general event-triggered framework for CBFs has been proposed in \cite{2022XiaoEventAuto}. 
It remains to be shown how this framework can be used for real CAVs with proper event-triggered communication, optimization, and control in the presence of measurement uncertainties, noise, and multiple safety constraints. 
In this paper, we address this question by adapting the framework in \cite{2022XiaoEventAuto} to CAVs and implementing it in a laboratory-scale test bed we have developed focusing on the problem of automated merging roadways and emulating CAVs through mobile robots.
Several optimal control methods have been developed for CAVs and implemented on actual mobile robots. CAV maneuvers have been demonstrated in a scaled environment with 2-3 CAVs in \cite{2019Karl}. Highway driving conditions have been created to validate CAV maneuvers in \cite{2019Hyldmar}. In \cite{2017Chalaki} and \cite{Beaver_2020} the unconstrained optimal control solution has been implemented on robots for the merging roadway scenario and the corridor scenario, respectively. To our knowledge, there is no general \emph{safety-guaranteed} control algorithm for CAVs implemented on real robots for merging roadways.

\begin{figure}[t]
    \centering
    \includegraphics[width=0.47\textwidth]{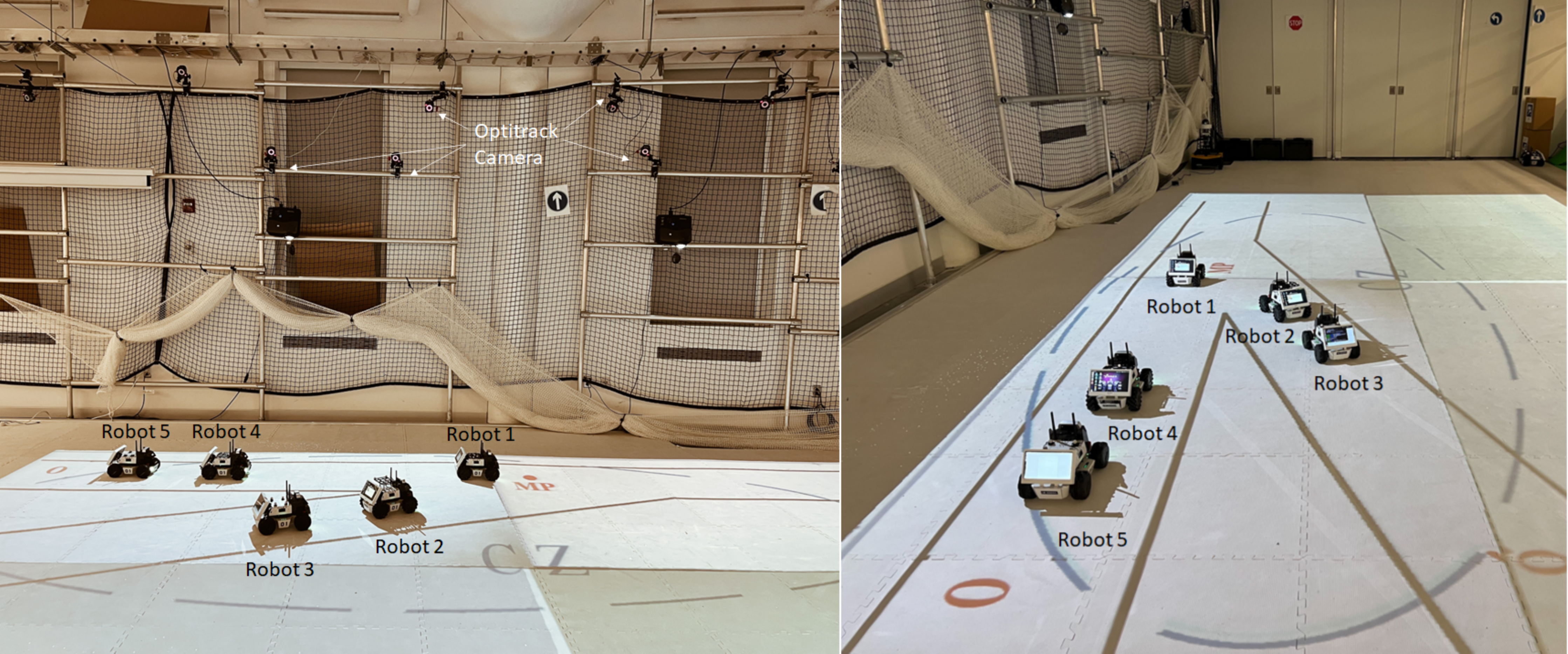}
    \caption{The merging experiment setup where the control zone (CZ) starts from point $O$ in the main road and point $O^{\prime}$ in the merging road and ends at the merging point $MP$.}
    \label{fig:realmergingroad}
\end{figure}

 \textit{Our contribution:} 
The contributions of the paper are as follows. $(i)$ We follow the approach presented in \cite{2022XiaoEventAuto} and introduce an \emph{event-triggered} framework for coordination and control of CAVs in merging roadways. The underlying concept behind the approach is to define \emph{events} associated with states reaching a certain bound, at which point the next QP instance is triggered. This approach provides manifold benefits, namely $(a)$ guaranteeing safety constraints are satisfied in the presence of measurement noise and model uncertainties which are inevitable in practical applications, as opposed to the time-driven scheme, and $(b)$ significant reduction in the number of QPs solved, thereby reducing the need for unnecessary communication among CAVs and overall computational overload. 
$(ii)$  In addition to validating the event-triggered framework and aforementioned benefits through simulation, we have also 
implemented it in a laboratory-scale test bed using mobile robots as shown in Fig \ref{fig:realmergingroad}. The test bed allows us to test a variety of cooperative control algorithms for CAVs under multiple safety constraints (such as rear-end safety and safe merging) in the presence of disturbances that cannot be captured in a simulated setting. Although the experiment was set up for a merging scenario as illustrated in Fig. \ref{fig:control_diagram}, it can be readily extended to other bottleneck points. In this test bed, all computation, including checking for bound violations and re-solving a QP to obtain a new control input, are performed on-board in a real-time manner.

The paper is organized as follows. In Section II, we provide an overview of the time-driven decentralized constrained optimal control for CAVs through the OCBF approach in merging roadways as an example of a conflict area, providing the motivation for the event-triggered approach. In section III, the event-triggered scheme is presented.
In section IV, implementation details, including the communication and motion control of the robots, are discussed. In section V, the implementation results are provided to demonstrate the effectiveness of the proposed event-triggered method in experiments with actual robots in terms of improved feasibility and guaranteed safety constraints. Additionally, extensive simulation results are provided to validate the proposed approach.

\section{Problem Formulation and Event-triggered Control}
\begin{figure}[ptbh]
\centering
\includegraphics[scale=0.6]{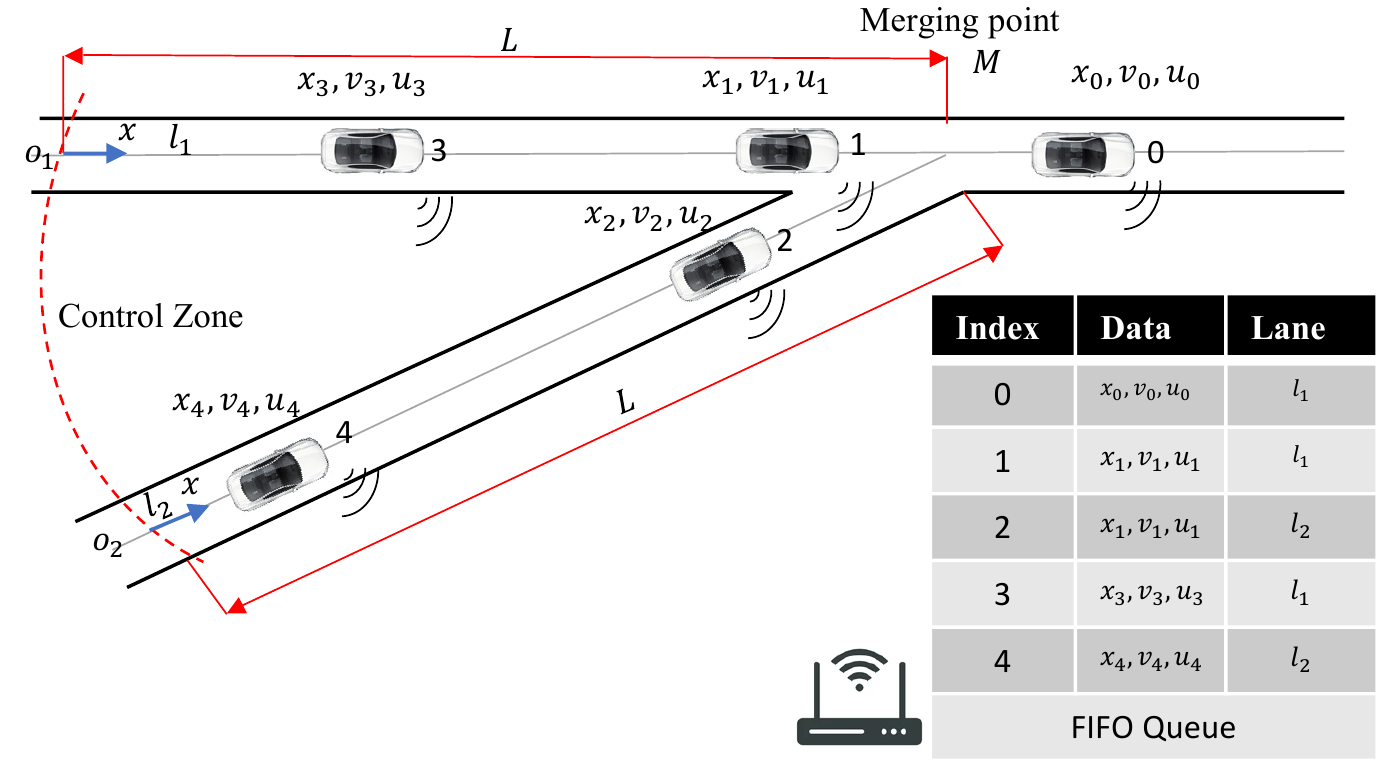} 
\caption{Illustration of cooperative control of CAVs at merging roadways.}%
\label{fig:merging}%
\end{figure}

In this section, we review the setting for CAVs whose motion is cooperatively controlled at conflict areas of a traffic network. This includes merging roads, signal-free intersections, roundabouts, and highway segments where lane change maneuvers take place. Although the framework can be used for all conflict areas, the focus on this paper is on merging roadways.  Following \cite{XIAO2021109333}, we define a \emph{Control Zone} (CZ) to be an area within which CAVs can communicate with each other or with a coordinator which is responsible for facilitating the exchange of information within this CZ. As an example, Fig. \ref{fig:merging} shows a conflict area due to vehicles merging from two single-lane roads and there is a single Merging Point (MP) which vehicles must cross from either road \cite{XIAO2021109333}. 
In such a setting, assuming all traffic consists of CAVs, a finite horizon constrained optimal control problem can be formulated aiming to determine trajectories that jointly minimize travel time and energy consumption through the CZ and guaranteeing safety constraints are always satisfied.

Let $S(t)$ be the set of First-In-First-Out (FIFO)-ordered indices of all CAVs located in the CZ at time $t$ along with the CAV that has just left the CZ (whose index is 0 as shown in Fig.\ref{fig:merging}). $N(t)$ is defined as the cardinality of $S(t)$. Thus, a CAV arriving at time $t$ is assigned $N(t)$ as its index. All CAV indices in the CZ $S(t)$ decrease by one when a CAV crosses the MP and the vehicle whose index is $-1$ is dropped.

The vehicle dynamics for each CAV $i\in S(t)$ take the following form
\begin{equation}
\left[
\begin{array}
[c]{c}%
\dot{x}_{i}(t)\\
\dot{v}_{i}(t)
\end{array}
\right]  =\left[
\begin{array}
[c]{c}%
v_{i}(t)\\
u_{i}(t)
\end{array}
\right],  \label{VehicleDynamics}%
\end{equation}
where $x_{i}(t)$ denotes the distance from the origin at which CAV $i$ arrives, $v_{i}(t)$, and $u_{i}(t)$ denotes the velocity and control input (acceleration) of CAV $i$, respectively.

\begin{remark}
The dynamics of CAVs and mobile robots in real-world applications are more complex than those in \eqref{VehicleDynamics}. In the test bed we have developed, the inadequacies of simple dynamics are compensated for through a local controller.
\end{remark}

Let $t_{i}^{0}$ and $t_{i}^{f}$ denote the time that CAV $i$ arrives at the origin and leaves the CZ at its exit point, respectively. Constraints for any conflict area can be listed as follows:
{\bf Constraint 1} (Safety constraints): Let $i_{p}$ denote the index of the CAV which physically immediately precedes $i$ in the CZ (if one is present). We require that the distance $z_{i,i_{p}}(t):=x_{i_{p}}(t)-x_{i}(t)$ be constrained by:
\begin{equation}
    z_{i,i_{p}}(t)\geq\varphi v_{i}(t)+\delta,\text{ \ }\forall t\in\lbrack
    t_{i}^{0},t_{i}^{f}] \label{Safety}%
\end{equation}
where $\varphi$ denotes the reaction time and $\delta$ is a given minimum safe distance. If we define $z_{i,i_{p}}$ to be the distance from the center of CAV $i$ to the center of CAV $i_{p}$, then $\delta$ depends on the length of these two CAVs.\\
{\bf Constraint 2} (Safe merging): Whenever a CAV crosses a MP, a lateral collision is possible and there must be adequate safe space for the CAV to avoid such collision, i.e.,
\begin{equation}
\label{SafeMerging}
z_{i,i_c}(t_{i}^{m})\geq\varphi v_{i}(t_{i}^{m})+\delta,
\end{equation}
where $i_c$ is the index of the CAV that may collide with CAV $i$ at the merging point $M$ as shown in Fig. \ref{fig:merging}. The determination of CAV $i_c$ depends on the policy adopted for sequencing CAVs through the CZ. 
For the merging roadway in Fig. \ref{fig:merging} under FIFO, we have $i_c=i-1$ and $t_i^m=t_i^f$ since the MP defines the exit from the CZ.\\
{\bf Constraint 3} (Vehicle limitations): Finally, there are constraints on the speed and acceleration for each $i\in S(t)$:
\begin{equation}
\label{VehicleConstraints1}%
\begin{aligned} v_{min} \leq v_i(t)\leq v_{max}, \forall t\in[t_i^0,t_i^f]\end{aligned} 
\end{equation}
\begin{equation}
\label{VehicleConstraints2}%
\begin{aligned} u_{{min}}\leq u_i(t)\leq u_{{max}}, \forall t\in[t_i^0,t_i^f] \end{aligned} 
\end{equation}
where $v_{\max}> 0$ and $v_{\min} \geq 0$ denote the maximum and minimum speed allowed
in the CZ, $u_{{\min}}<0$ and $u_{\max}>0$ denote the minimum and maximum
control, respectively.

\textbf{Decentralized Optimal Control Problem formulation.} Our goal is to determine a control law jointly minimizing the travel time and energy consumption subject to constraints 1-3 for each $i \in S(t)$ governed by the dynamics (\ref{VehicleDynamics}). Expressing energy through $\frac{1}{2}u_i^2(t)$ and normalizing travel time and energy, we use the weight $\alpha_i\in[0,1]$ to construct
a convex combination as follows:
\begin{equation}\label{eqn:energyobja_m}
\begin{aligned}\min_{u_{i}(t),t_i^f} J_i(u_i(t),t_i^f)= \int_{t_i^0}^{t_i^f}\left(\alpha_i + \frac{(1-\alpha_i)\frac{1}{2}u_i^2(t)}{\frac{1}{2}\max \{u_{max}^2, u_{min}^2\}}\right)dt \end{aligned}.
\end{equation}
 Letting $\beta_i:=\frac{\alpha_i\max\{u_{max}^{2},u_{min}^{2}\}}{2(1-\alpha_i)}$, we obtain a simplified form: 
\begin{equation}\label{eqn:energyobja}
\min_{u_{i}(t),t_i^f}J_{i}(u_{i}(t),t_i^f):=\beta_i(t_{i}^{f}-t_{i}^{0})+\int_{t_{i}^{0}%
}^{t_{i}^{f}}\frac{1}{2}u_{i}^{2}(t)dt,
\end{equation}
where $\beta_i \geq0$ is an adjustable weight to penalize
travel time relative to the energy cost of CAV $i$. Note that the solution is \emph{decentralized} in the sense that CAV $i$ requires information only from CAVs $i_p$ and $i_c$ required in (\ref{Safety}) and (\ref{SafeMerging}).

\textbf{The OCBF approach}. Firstly, we derive the CBFs that ensure the constraints (\ref{Safety}), (\ref{SafeMerging}), and (\ref{VehicleConstraints1}) are always satisfied, subject to the vehicle dynamics (\ref{VehicleDynamics}) by defining $f(\boldsymbol{x}_i(t))=[v_i(t),0]^T$ and $g(\boldsymbol{x}_i(t))=[0,1]^T$. Each of these constraints can be easily written in the form of $b_q(\boldsymbol{x}(t)) \geq 0$, $q \in \lbrace  1,...,n \rbrace$ where $n$ stands for the number of constraints and $\boldsymbol{x}(t)=[\boldsymbol{x}_1(t),\boldsymbol{x}_2(t),...,\boldsymbol{x}_{N(t)}(t)]$. The CBF method (details provided in \cite{Xiao_03}) maps a constraint $b_q(\boldsymbol{x}(t)) \geq 0$ onto a new constraint which is \emph{linear} in the control input and takes the general form 
\begin{equation} \label{cbf_condition}
L_fb_q(\boldsymbol{x}(t))+L_gb_q(\boldsymbol{x}(t))u_i(t)+\gamma( b_q(\boldsymbol{x}(t))) \geq 0.
\end{equation}

To obtain the CBF constraint for the safety constraint \eqref{Safety} we set $b_1(\boldsymbol{x}_i(t),\boldsymbol{x}_{i_p}(t))=z_{i,i_{p}}(t)-\varphi v_{i}(t)-\delta=x_{i_p}(t)-x_i(t)-\varphi v_i(t)-\delta$ and since $b_1(\boldsymbol{x}_i(t),\boldsymbol{x}_{i_p}(t))$ is differentiable,
the CBF constraint for (\ref{Safety}) is
\begin{equation}\label{CBF1}\small
\underbrace{v_{i_p}(t)-v_i(t)}_{L_fb_1(\boldsymbol{x}_i(t),\boldsymbol{x}_{i_p}(t))}+\underbrace{-\varphi}_{L_gb_1(\boldsymbol{x}_i(t))} u_i(t)+\underbrace{k_1(z_{i,i_p}(t)-\varphi v_i(t) - \delta)}_{\gamma_1(b_1(\boldsymbol{x}_i(t),\boldsymbol{x}_{i_p}(t)))} \geq 0,
\end{equation}
where the class-$\mathcal{K}$ function $\gamma_1(\boldsymbol{x})$ is chosen here to be linear.

Deriving the CBF constraint for the safe merging constraint (\ref{SafeMerging}) poses a technical challenge due to the fact that it only applies at a certain time $t_i^{m}$, whereas a CBF is required to be in a continuously differentiable form. To tackle this problem, by following the technique used in \cite{Xiao_03},
we replace $\phi$ in (\ref{SafeMerging}) by $\Phi(x_i(t))=\varphi  \frac{x_i(t)}{L} $,
where $L$ is the length of the road traveled by the CAV from the CZ entry to the MP,
and use $\label{b2} 
    b_2(\textbf{x}_i(t),\textbf{x}_{i_c}(t))= x_{i_c}(t)-x_i(t)-\Phi(x_i(t)) v_i(t)-\delta$
so that the CBF constraint for safe merging \eqref{SafeMerging} becomes

\begin{align}\small \label{CBF2}
&\underbrace{v_{i_c}(t)-v_i(t)-\frac{\varphi}{L}v_i^2(t)}_{L_fb_2(\boldsymbol{x}_i(t),\boldsymbol{x}_{i_c}(t))}+\underbrace{-\varphi \frac{x_i(t)}{L}}_{L_gb_2(\boldsymbol{x}_i(t))}u_i(t)+\nonumber \\ &\underbrace{k_2(z_{i,i_c}(t)-\varphi  \frac{x_i(t)}{L} v_i(t)}_{\gamma_2(b_2(\boldsymbol{x}_i(t),\boldsymbol{x}_{i_c}(t)))} \geq 0.
\end{align}

The speed constraints in \eqref{VehicleConstraints1} are also easily transformed into CBF constraints by defining $b_3(\boldsymbol{x}_i(t))=v_{max}-v_i(t)$ and $b_4(\boldsymbol{x}_i(t))=v_i(t)-v_{min}$. This yields:
\begin{align} \label{CBF3-4}
\underbrace{-1}_{L_gb_3(\boldsymbol{x}_i(t))}u_i(t)+\underbrace{k_3(v_{max}-v_i(t))}_{\gamma_3(b_3(\boldsymbol{x}_i(t)))} \geq 0 \nonumber, \\
\underbrace{1}_{L_gb_4(\boldsymbol{x}_i(t))}u_i(t)+\underbrace{k_4(v_i(t)-v_{min})}_{\gamma_4(b_4(\boldsymbol{x}_i(t)))} \geq 0,
\end{align}
for the maximum and minimum velocity constraints, respectively.

As a last step in the OCBF approach, we use a Control Lyapunov Function (CLF) associated with tracking the CAV speed to a desired value $v_{i}^{ref}(t)$ setting $V(\boldsymbol{x}_i(t))=(v_i(t)-v_{i}^{ref}(t))^2$ and express the CLF constraint as follows:
\begin{equation}\label{CLF_constraint}
L_fV(\boldsymbol{x}_i(t))+L_gV(\boldsymbol{x}_i(t))\boldsymbol{u}_i(t)+ c_3 V(\boldsymbol{x}_i(t))\leq e_i(t),
\end{equation}
where $e_i(t)$ makes this a soft constraint.

Finally, we can formulate the OCBF problem as follows:
\begin{equation}\label{QP-OCBF}\small
\min_{u_i(t),e_i(t)}J_i(u_i(t),e_i(t)):=\int_{t_i^0}^{t_i^f}\big[\frac{1}{2}(u_i(t)-u_{i}^{ref}(t))^2+\lambda e^2_i(t)\big]dt
\end{equation}
subject to vehicle dynamics (\ref{VehicleDynamics}), the CBF constraints (\ref{CBF1}), (\ref{CBF2}), \eqref{CBF3-4}, and CLF constraint (\ref{CLF_constraint}).
In this approach, $(i)$ we solve the unconstrained optimal control problem in \eqref{eqn:energyobja} which yields $u_i^{ref}$, $(ii)$ the
resulting $u_i^{ref}$ is optimally tracked such that constraints including CBF constraints (\ref{CBF1}), (\ref{CBF2}), \eqref{CBF3-4} satisfied, and $(iii)$ this tracking optimal control problem is efficiently solved by discretizing time and solving a simple QP at each discrete time step.

As already mentioned, a common way to solve this dynamic optimization problem is to discretize $[t_i^0,t_i^f]$ into intervals $[t_i^0,t_i^0+\Delta],...,[t_i^0+k\Delta,t_i^0+(k+1)\Delta],...$ with equal length $\Delta$ and solving (\ref{QP-OCBF}) over each time interval. The decision variables $u_{i,k}=u_i(t_{i,k})$ and $e_{i,k}=e_i(t_{i,k})$ are assumed to be constant on each interval and can be easily calculated at time $t_{i,k}=t_i^0+k\Delta$ through solving a QP at each time step:
\begin{align} \label{QP}
\min_{u_{i,k},e_{i,k}}&[ \frac{1}{2}(u_{i,k}-u_i^{ref}(t_{i,k}))^2+\lambda e_{i,k}^{2}]
\end{align} 
subject to the constraints (\ref{CBF1}), (\ref{CBF2}), \eqref{CBF3-4}, (\ref{CLF_constraint}), and \eqref{VehicleConstraints2} where all constraints are linear in the decision variables. We refer to this as the \emph{time-driven} approach. As pointed out earlier, the main problem with this approach is that there is no guarantee for the feasibility of each CBF-based QP, as it requires a small enough discretization time which is not always possible to achieve. Also it is worth mentioning that requiring synchronization among all CAVs can be difficult to impose in real-world applications.

\section{Event-triggered control}
In this paper, we adopt an \emph{event-triggering} scheme whereby a QP (with its associated CBF constraints) is solved when one of three possible events (as defined next) is detected. We will show that this provides a guarantee for the satisfaction of the safety constraints which cannot be offered by the time-driven approach described earlier. As introduced in \cite{2022XiaoEventAuto}, the key idea is to ensure that the safety constraints are satisfied while the state remains within some bounds and define events which coincide with the state reaching these bounds, at which point the next instance of the QP in (\ref{QP}) is triggered.

Let $t_{i,k}$, $k=1,2,...$, be the time instants when CAV $i$ solves the QP in (\ref{QP}). Our goal is to guarantee that the state trajectory does not violate any safety constraints within any time interval $(t_{i,k},t_{i,k+1}]$ where $t_{i,k+1}$ is the next time instant when the QP is solved. We define a subset of the state space of CAV $i$ at time $t_{i,k}$ such that:
\begin{equation} \label{bound}
\boldsymbol{x}_i(t_{i,k})-\boldsymbol{s}_i \leq \boldsymbol{x}_i(t) \leq \boldsymbol{x}_i(t_{i,k})+\boldsymbol{s}_i,
\end{equation}
where $\boldsymbol{s_i} \in \mathbb{R}_{>0}^2$ is a parameter vector whose choice will be discussed later. 
We denote the set of states of CAV $i$ that satisfy \eqref{bound} at time $t_{i,k}$ by 
\begin{equation} \label{event bound}
S_i(t_{i,k}) = \bigl\{\boldsymbol{y}_i \in \textbf{X}: \boldsymbol{x}_i(t_{i,k})-\boldsymbol{s}_i \leq \boldsymbol{y}_i \leq \boldsymbol{x}_i(t_{i,k})+\boldsymbol{s}_i\bigl\}.
\end{equation} 
In addition, let $C_{i,1}$ be the feasible set of our original constraints \eqref{Safety},\eqref{SafeMerging} and \eqref{VehicleConstraints1} defined as
\begin{equation} \label{event:ci1}
    C_{i,1}:=\Bigl\{  \mathbf{x}_i\in \mathbf{X}: ~b_q(\mathbf{x})\geq 0, \ q \in \lbrace 1,2,3,4 \rbrace \Bigl\},
\end{equation}
Next, we seek a bound and a control law that satisfies the safety constraints within this bound. This can be accomplished by considering the minimum value of each component of \eqref{cbf_condition}, where $q \in \lbrace1,2,3,4\rbrace $,  as shown next.

Let us start with the first term in \eqref{cbf_condition}, $L_fb_q(\boldsymbol{x}(t))$, rewritten as $L_fb_q(\boldsymbol{y}_i(t),\boldsymbol{y}_r(t))$ with $\boldsymbol{y}_i(t)$ as in (\ref{event bound}) and $r$ in $\boldsymbol{y}_r(t)$ stands for \enquote{relevant} CAVs affecting the constraint of $i$ 
(i.e., $r \in \lbrace i_p, i_c \rbrace$ in Fig. \ref{fig:merging}).
Let $b^{min}_{q,f_i}(t_{i,k})$ be the minimum possible value of the term $L_fb_q(\boldsymbol{x}(t))$ 
over the time interval $(t_{i,k},t_{i,k+1}]$ for each $q \in \lbrace1,2,3,4\rbrace$ over the set $\Bar{S_i}({t_{i,k}}) \cap \Bar{S_r}({t_{i,k}})$:
\begin{equation}\label{minfi}
b^{min}_{q,f_i}(t_{i,k})=\displaystyle\min_{\boldsymbol{y}_i \in \Bar{S}_i({t_{i,k}}) \atop \boldsymbol{y}_r \in \Bar{S}_r({t_{i,k}})}L_fb_q(\boldsymbol{y}_i(t),\boldsymbol{y}_r(t)),
\end{equation} 
where $\Bar{S}_i({t_{i,k}})$ is defined as follows:
\begin{equation}
    \Bar{S}_i({t_{i,k}}):=\bigl\{\boldsymbol{y}_i \in C_{i,1} \cap S_i(t_{i,k})\bigl\}
\end{equation}

Similarly, we can define the minimum value of the third term in \eqref{cbf_condition}:
\begin{equation}\label{mingammai}
b^{min}_{\gamma_q}(t_{i,k})=\displaystyle\min_{\boldsymbol{y}_i \in \Bar{S}_i({t_{i,k}}) \atop \boldsymbol{y}_r \in \Bar{S}_r({t_{i,k}})} \gamma_q(\boldsymbol{y}_i(t),\boldsymbol{y}_r(t)).
\end{equation}

For the second term in \eqref{cbf_condition}, note that $L_gb_q(\boldsymbol{x}_i)$ is a constant for $ q \in\lbrace1,3,4\rbrace $, therefore there is no need for any minimization. However, $L_gb_2(\boldsymbol{x}_i)$ in (\ref{CBF2}) is state dependent and needs to be considered for the minimization. Since $x_i(t) \ge 0$, note that $L_gb_2(\boldsymbol{x}_i)=-\varphi \frac{x_i(t)}{L}$ is always negative, therefore, we can determine the limit value $b^{min}_{2,g_i}(t_{i,k}) \in \mathbb{R}, $ as follows:
\begin{eqnarray}\label{mingi} \small
b^{min}_{2,g_i}(t_{i,k})=\begin{cases}
\displaystyle\min_{\boldsymbol{y}_i \in \Bar{S}_i({t_{i,k}}) \atop \boldsymbol{y}_r \in \Bar{S}_r({t_{i,k}})}L_gb_2(\boldsymbol{x}_i(t)), \  \textnormal{if}\  u_{i,k} \geq 0\\
\\
\displaystyle\max_{\boldsymbol{y}_i \in \Bar{S}_i({t_{i,k}}) \atop \boldsymbol{y}_r \in \Bar{S}_r({t_{i,k}})}L_gb_2(\boldsymbol{x}_i(t)), \ \ \  \textnormal{otherwise},
\end{cases}
\end{eqnarray}
where the sign of $u_{i,k},  \ i \in S(t_{i,k})$ can be determined by simply solving the CBF-based QP  \eqref{QP-OCBF} at time $t_{i,k}$.

Thus, the condition that can guarantee the satisfaction of \eqref{CBF1},\eqref{CBF2} and \eqref{CBF3-4} in the time interval $\left(t_{i,k},t_{i,k+1}\right]$ is given by
\begin{equation} \label{minCBF}
b^{min}_{q,f_i}(t_{i,k})+b^{min}_{q,g_i}(t_{i,k})u_{i,k}+b^{min}_{\gamma_q}(t_{i,k})\geq 0,
\end{equation}
for $q\in\lbrace1,2,3,4\rbrace$. In order to apply this condition to the QP \eqref{QP-OCBF}, we just replace \eqref{cbf_condition} by \eqref{minCBF} as follows:
\begin{align} \label{eq:QPtk}
\min_{u_{i,k},e_{i,k}}&[ \frac{1}{2}(u_{i,k}-u_i^{ref}(t_{i,k}))^2+\lambda e_{i,k}^{2}]\nonumber\\
 &\textnormal{s.t.} \ \  \eqref{CLF_constraint},\eqref{minCBF}, \eqref{VehicleConstraints2}
\end{align} 
It is important to note that each instance of the QP \eqref{eq:QPtk} is now triggered by one of the following three events where $k =1,2,\ldots$ is an event (rather than time step) counter:\\ 
\textbf{Event 1:} the state measurement of CAV $i$ reaches the boundary of $S_i(t_{i,k-1})$.
\\
\textbf{Event 2:} the state measurement of CAV $i_p$ reaches the boundary of $S_{i_p}(t_{i,k-1})$ (if $i_p$ exists for CAV $i$).
\\
\textbf{Event 3:} the state measurement of CAV $i_c$ reaches the boundary of $S_{i_c}(t_{i,k-1})$  where $i_c$ is the index of the CAV that may collide with $i$ in (\ref{SafeMerging}), e.g., $i_c=i-1 \neq i_p$ in the merging problem case, if such a CAV exists.

 The state measurements are obtained from external sensors which introduce noise in the state values. Thus, we denote the state measurements of CAV $i$ by $\Tilde{\mathbf{x}}_i(t)$ defined as follows:
 \begin{equation}
      \Tilde{\boldsymbol{x}}_i(t)= \boldsymbol{x}_i(t)+ \boldsymbol{w}(t)
 \end{equation}
where $\boldsymbol{w(t)}$ is a random but bounded noise term. As a result, $t_{i,k},k=1,2,...$ is unknown in advance but can be determined by CAV $i$ through:
\begin{align} \label{events}
t_{i,k}=\min \Big\{ t>t_{i,k-1}:\vert\Tilde{\boldsymbol{x}}_i(t)-\Tilde{\boldsymbol{x}}_i(t_{i,k-1})\vert=\boldsymbol{s}_i \\ \nonumber
\text{or} \ \ \vert\Tilde{\boldsymbol{x}}_{i_p}(t)-\Tilde{\boldsymbol{x}}_{i_p}(t_{i,k-1})\vert=\boldsymbol{s}_{i_p} \\ \nonumber 
\text{or} \ \ \vert\Tilde{\boldsymbol{x}}_{i_c}(t)-\Tilde{\boldsymbol{x}}_{i_c}(t_{i,k-1})\vert=\boldsymbol{s}_{i_c}\Big\},
\end{align}
where $t_{i,1}=0$. 

The following theorem formalizes our analysis by showing that if new constraints of the general form \eqref{minCBF} holds, then our original CBF constraints \eqref{CBF1},\eqref{CBF2} and \eqref{CBF3-4} also hold. The proof follows the same lines as that of a more general theorem in \cite{2022XiaoEventAuto} and, therefore, is omitted.

\begin{theorem}\label{as:1} Given a CBF $b_q(\mathbf{x(t)})$ with relative degree one, let $t_{i,k+1}$, $k=1,2,\ldots$ be determined by \eqref{events} with $t_{i,1}=0$ and $b^{min}_{q,f_i}(t_{i,k})$, $b^{min}_{\gamma_q}(t_{i,k})$, $b^{min}_{q,g_i}(t_{i,k})$  for $q=\{1,2,3,4\}$ obtained through \eqref{minfi}, \eqref{mingammai}, and \eqref{mingi}. Then, any control input $u_{i,k}$ that satisfies \eqref{minCBF} for all $q \in [1,2,3,4]$  within the time interval $[t_{i,k},t_{i,k+1})$ renders the set $C_{i,1}$ forward invariant for the dynamic system defined in (\ref{VehicleDynamics}).
\end{theorem}

\begin{remark}:
Expressing \eqref{minCBF} in terms of the minimum value of each component separately may become overly conservative if each minimum value corresponds to different points in the decision variable space. Therefore, an alternative approach is to calculate the minimum value of the whole term. 
\end{remark}
The selection of the parameters $\mathbf{s}_i$ captures the trade-off between \emph{computational cost} and \emph{conservativeness}: the larger the value of each component of $\mathbf{s}_i$ is, the smaller the number of events that trigger instances of the QPs becomes, thus reducing the total computational cost. At the same time, the control law must satisfy the safety constraints over a longer time interval as we take the minimum values in \eqref{minfi}-\eqref{mingi}, hence rendering the approach more conservative. The choice of parameter $\mathbf{s}_i$ can also render our algorithm more robust to measurement uncertainties and noise,
hence making it useful for real-world implementation. To account for measurement uncertainties, the value of each bound should be greater than or equal to the measurement noise, i.e., the inequality $\mathbf{s}_i \geq \sup \mathbf{w}(t)$ should hold componentwise. 

\section{Test bed for Coordinated Merging}
\begin{figure*}
    \centering
    \includegraphics[scale=.55]{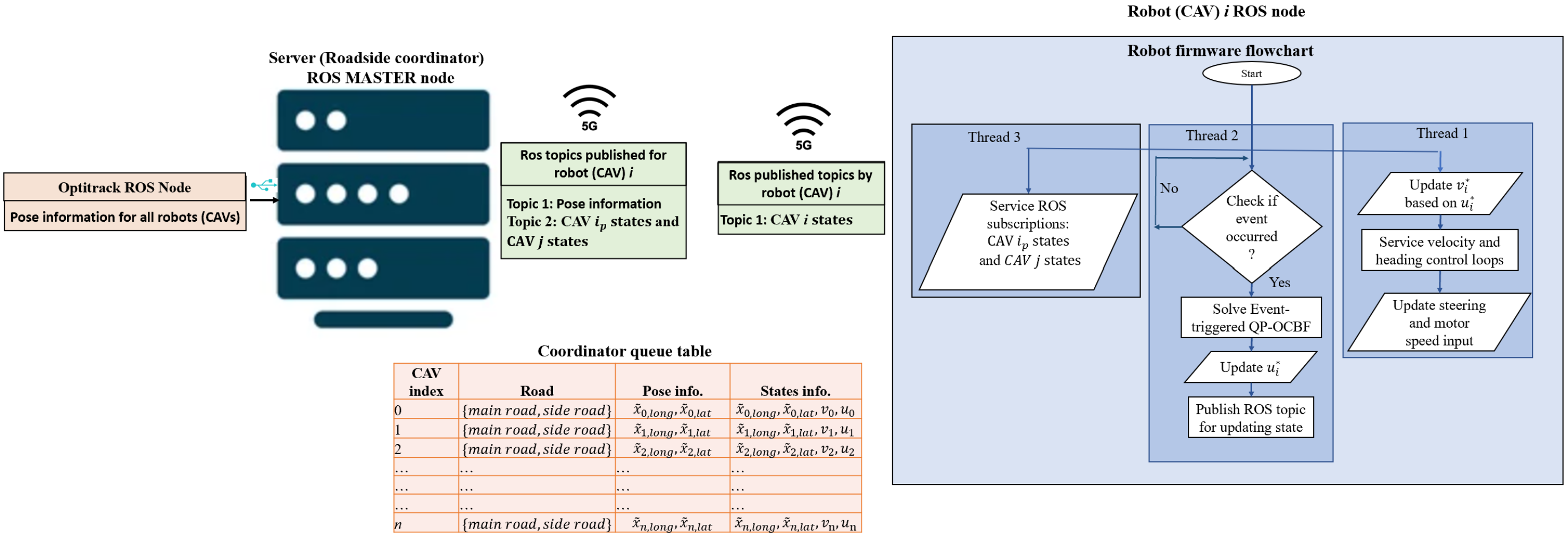}
    \caption{Illustration of the testbed implementation.}
    \label{fig:control_diagram}
\end{figure*}

The event-triggered coordinated merging process described in the last section was implemented in a lab environment using mobile robots as CAVs, while the coordinator was implemented on a laptop located in the vicinity of the robots. The overall implementation can be divided into two main parts: $(i)$ communication and $(ii)$ control. A block-level illustration of the architecture of the robots firmware is presented in Fig. \ref{fig:control_diagram} and provides the basis for designing similar test beds for related research.

\textbf{Communication}. The OCBF approach only requires vehicle-to-infrastructure (V2I) communication, whereby the coordinator is responsible for exchanging information among robots. The communication was implemented using a 5G wireless LAN where the messages were exchanged using ROS Topics. In an event-triggered scheme, frequent communication is generally not needed, since it occurs only when an event is triggered. The events are state-dependent, thus each robot in the CZ requires its position information. Since the robots are not equipped with GPS, the position and orientation (POSE) information was obtained using the Optitrack localization system by the coordinator using ROS topics over Ethernet as shown in Fig. \ref{fig:control_diagram}. The coordinator uses the localization information to index the robots upon arrival in the CZ for constructing the queue table based on the FIFO scheme as illustrated in Fig. \ref{fig:merging}) and transmits the indices to them. Additionally, the POSE information of robots is transmitted to themselves periodically while in the CZ.

Each robot is responsible for checking its own state to detect any violation in its state bounds in (\ref{event bound}). When such an event occurs, the robot updates its control input by re-solving the QP in \eqref{eq:QPtk} and informs the coordinator with its newly obtained state (i.e., velocity and position) by publishing a ROS topic. The structure of data packets sent by any robot with index $i$ is described using the vector $(i, \Tilde{x}_{i,long}, \Tilde{x}_{i,lat}, \Tilde{v}_{i})^T$ where $\Tilde{x}_{i,long}$ and  $\Tilde{x}_{i,lat}$ denote the position along and orthogonal to the direction of the main road, respectively and $\Tilde{v}_i$ denotes the linear speed along the direction of the road of robot $i$. It then becomes the responsibility of the coordinator to provide this information to the relevant robots (i.e., those that might be affected). There can be at most two robots that can be relevant to any robot $i$, the one which is immediately following it in the same lane and the robot that will merge after it from the other lane. Note that a triggered event due to an update in robot $i$'s state can affect only robots $l>i$. The data are published as ROS topics by the coordinator separately for each one (or two) of the robot(s) in the form $(\Tilde{x}_{{i_p},long}, \Tilde{x}_{{i_p},lat}, \Tilde{v}_{i_p}, \Tilde{x}_{i_c,long}, \Tilde{x}_{i_c,lat}, \Tilde{v}_{i_c})^T$  where  $\Tilde{x}_{{i_p},long}$ and $\Tilde{x}_{{i_p},lat}$ denote the lateral and longitudinal position and $\Tilde{v}_{i_p}$ denotes the linear speed of the preceding robot $i_p$, and $\Tilde{x}_{{i_c},lat}$ and  $\Tilde{x}_{{i_c},long}$ denote the lateral and longitudinal position and $\Tilde{v}_{i_c}$ denotes the linear speed of robot $i_c$ that will merge before the robot. The robots subscribe to the topics published for them using the index as the identifier. As can be seen from Fig. \ref{fig:control_diagram}, the robot firmware contains a dedicated thread to handle the ROS messages from the coordinator. Finally, the notified robots decide whether they need to re-solve their QP or maintain their control input until the next triggering event by checking for a potential bound violation. As a result, they either need to re-solve their QP, and send their updated state to the coordinator or they carry on with their current control input until the next event.

\textbf{Control}.
The event-triggered QP in \eqref{eq:QPtk} for any robot $i$ returns the acceleration ${u}^*_{i,k}$, which has to be controlled along with the trajectory of every robot in the CZ to prevent them from deviating from the road. The two control objectives are achieved using feedback control. The linear acceleration for the robots cannot be directly controlled, hence the desired acceleration  ${u}^*_{i,k}$ for any robot $i$ is mapped onto its desired linear velocity $v^*_{i}(t) \ \forall t \in [t_{i,k},t_{i,k+1}) \ \textnormal{and} \ \forall k$ using the mapping below. 
\begin{equation}
\label{v_ref}
    v^*_i(t+\Delta T) = {u}^*_{i,k}\Delta T+v^*_i(t) \\\ \ \  t \in [t_{i,k},t_{i,k+1}), \ \forall k
\end{equation}
where $\Delta T$ is the period of the control loop execution. In order to track the acceleration $u^*_{i,k}$ of robot $i$ the value of $\Delta T$ needs to be generally chosen to be five times smaller than the inter-event time $(t_{i,k+1} - t_{i,k})$. However, since events occur asynchronously, $\Delta T$ is chosen to be a fifth of the sensor sampling rate at which the measurements are received, as the event-triggered approach solves a more conservative version of the time-driven OCBF algorithm. The velocity of the robot at any time $t$ is computed using the on-board wheel encoders and used to control the velocity by manipulating the motor speed using a PID controller.

It is desired that the robot follow the center of the lane while in the CZ, which is programmed as a set of way-points. The deviation from the center of the lane for any robot $i$ at time $t$ is denoted by $\Tilde{e}_i = \Tilde{x}_{i,lat}(t)-L_c$, where $L_c$ is the center of the road. The error can be negated by manipulating the steering angle using a PID controller whose gain is set through trial and error. Due to the nature of event-triggered control, two threads are dedicated in every robot to achieve the control tasks where the first thread is responsible for controlling the motion of the robot (executed every $\Delta T$ seconds) and the other for the event-triggered control (executed asynchronously upon the occurrence of any event as shown in Fig. \ref{fig:control_diagram}).


\section{Numerical Results}
All algorithms in this paper were implemented using \textsc{Python}. We used \textsc{QP} from the \textsc{cvxopt} library for solving \eqref{eq:QPtk}. Besides that, \textsc{lingprog} from the \textsc{scipy} library has been used to solve the linear program in \eqref{minfi}, \eqref{mingammai} and \eqref{mingi}, \textsc{fsolve} was used for a nonlinear optimization problem arising when \eqref{minfi} and \eqref{mingammai} become nonlinear respectively, and \textsc{solve\_ivp} was used to solve for the vehicle dynamics \eqref{VehicleDynamics}. Note that for results obtained in the test bed, all algorithms were executed on board each robot.

We have considered the merging problem shown in Fig. \ref{fig:realmergingroad} where robots arrive to the predefined CZ according to Poisson arrival processes with a given arrival rate. The initial speed $v_{i}(t_{i}^{0})$ is also randomly generated with a uniform distribution over $[0.1 \textnormal{m/s}, 1\textnormal{m/s}]$ at the origins $O$ and $O^{\prime}$, respectively. The
parameters for \eqref{QP-OCBF} and \eqref{eq:QPtk}
 are: $L = 3.04\textnormal{m}, \varphi = 0.18\textnormal{s}, \delta = 15\textnormal{cm}, u_{max} = 2 \ \textnormal{m/s}^2, u_{min} = -2 \ \textnormal{m/s}^2, v_{max} = 1\textnormal{m/s}, v_{min} = 0\textnormal{m/s},  k_1=k_2=k_3=k_4=1,  \lambda= 10$. The sampling rate of the localization data is $30$Hz, sufficiently high to avoid missing any triggering event (in general, a proper sampling rate can always be calculated given the CAV specifications, i.e., bounds on velocity and acceleration).
 The control update period for the time-driven control is $\Delta =0.05$s and $\Delta T=0.25$s. For the event-triggered control, we let the bounds $S=[s_x,s_v]$ be the same for all robots in the network and set to $[0.25,0.05]$.
 
 \subsection{Simulation results}
We first used a \textsc{Matlab}-based simulation environment using the python modules developed for actual implementation in the test bed. It is worth noting that some of the parameters for simulation and implementation are different as we had used mobile robots and did the implementations in a lab environment. Moreover during simulation, we included the computation of a more realistic energy consumption model \cite{kamal2012model} to supplement the simple surrogate $L_2$-norm ($u^2$) model in our analysis:
$
f_v(t)=f_{cruise}(t)+f_{accel}(t), 
f_{cruise}(t)= \omega_0+\omega_1v_i(t)+\omega_2v^2_i(t)+\omega_3v^3_i(t), 
f_{accel}(t)=(r_0+r_1v_i(t)+r_2v^2_i(t))u_i(t)
$
where we used typical values for parameters $\omega_1,\omega_2,\omega_3,r_0,r_1$ and, $r_2$ as reported in \cite{kamal2012model}. Note that in the test bed implementation, the energy term used in \eqref{eqn:energyobja_m} does not signify the energy dissipation (as it is difficult to measure the energy expended by the robots); rather, this term captures the control effort (i.e. acceleration/deceleration) as it is related to the energy consumed.  

Table I summarizes our results from 8 separate simulations corresponding to both the event-triggered and time-driven methods under the same conditions with different values for the relative weight of energy vs time as shown in the table. We observe that by using the event-triggered approach we are able to significantly reduce the number of infeasible QP cases (up to $86\%$) compared to the time-driven approach. At the same time, the overall number of instances when a QP needs to be solved has also decreased up to $50\%$ in the event-triggered case. Note that the large majority of infeasibilities are due to holding the control input (i.e. acceleration/develeration) constant over the sampling period, which can invalidate the forward invariance property of CBFs over the entire time interval. These infeasible cases were eliminated by the event-triggering approach. However, another source of infeasibility is due to  conflicts that may arise between the CBF constraints and the control bounds in a QP. This cannot be remedied through the event-triggered approach; it can, however, be dealt with by the introduction of a sufficient condition that guarantees no such conflict, as described in \cite{XIAO2022inf}. 
\begin{table*}\scriptsize
        \centering
        \begin{tabular}{|c|c|c|c|}
            \cline{1-4}
             &Item & \multicolumn{1}{|c|}{Event triggered} &Time Triggered\\
        \hline  
        \multirow{4}{*}{\makecell{$\alpha=0.1$ }}  & Ave. Travel time & 15.53  & 15.01 \\
        \cline{2-4}
        & Ave. $\frac{1}{2} u^2$  & 5.16 & 3.18\\
        \cline{2-4}
        & Ave. Fuel consumption  & 31.04 & 31.61\\
        \cline{2-4}
        &Computation load (Num of QPs solved) &  34\% (12168) & 100\% (35443) \\
        \cline{2-4}
        & Num of infeasible cases   & 43 & 315\\
        \hline
        \multirow{4}{*}{\makecell{$\alpha=0.25$ }}  & Ave. Travel time & 15.53  & 15.01 \\
        \cline{2-4}
        & Ave. $\frac{1}{2} u^2$&  14.25 & 13.34\\
        \cline{2-4}
        & Ave. Fuel consumption  & 51.42 & 55.81 \\
        \cline{2-4}
        &Computation load (Num of QPs solved) & 48\% (13707)  & 100\% (28200)\\
        \cline{2-4}
        & Num of infeasible cases   & 28 & 341 \\
        \hline
        \multirow{4}{*}{\makecell{$\alpha=0.4$ }}  & Ave. Travel time & 15.53  & 15.01 \\
        \cline{2-4}
        & Ave. $\frac{1}{2} u^2$&  18.22 & 17.67\\
        \cline{2-4}
        & Ave. Fuel consumption & 52.42 &  56.5\\
        \cline{2-4}
        &Computation load (Num of QPs solved)    & 49\% (13573) &  100\% (27412)\\
        \cline{2-4}
        & Num of infeasible cases   & 25 & 321 \\
        \hline       
        \multirow{4}{*}{\makecell{$\alpha=0.5$ }}  & Ave. Travel time & 15.17 &  14.63 \\
        \cline{2-4}
        & Ave. $\frac{1}{2} u^2$ & 24.93 & 25.08\\
        \cline{2-4}
        & Ave. Fuel consumption &  53.21 & 56.93 \\
        \cline{2-4}
        &Computation load (Num of QPs solved) & 50\% (13415) & 100\% (26726) \\
        \cline{2-4}
        & Num of infeasible cases   & 20 & 341\\
        \hline        
        \end{tabular}
        \caption{CAV metrics under event-driven and time-driven control. }
        \label{Table I}
\end{table*}

 \subsection{Test Bed Implementation Results}
\begin{figure}
   \centering
    \includegraphics[width=0.4\textwidth]{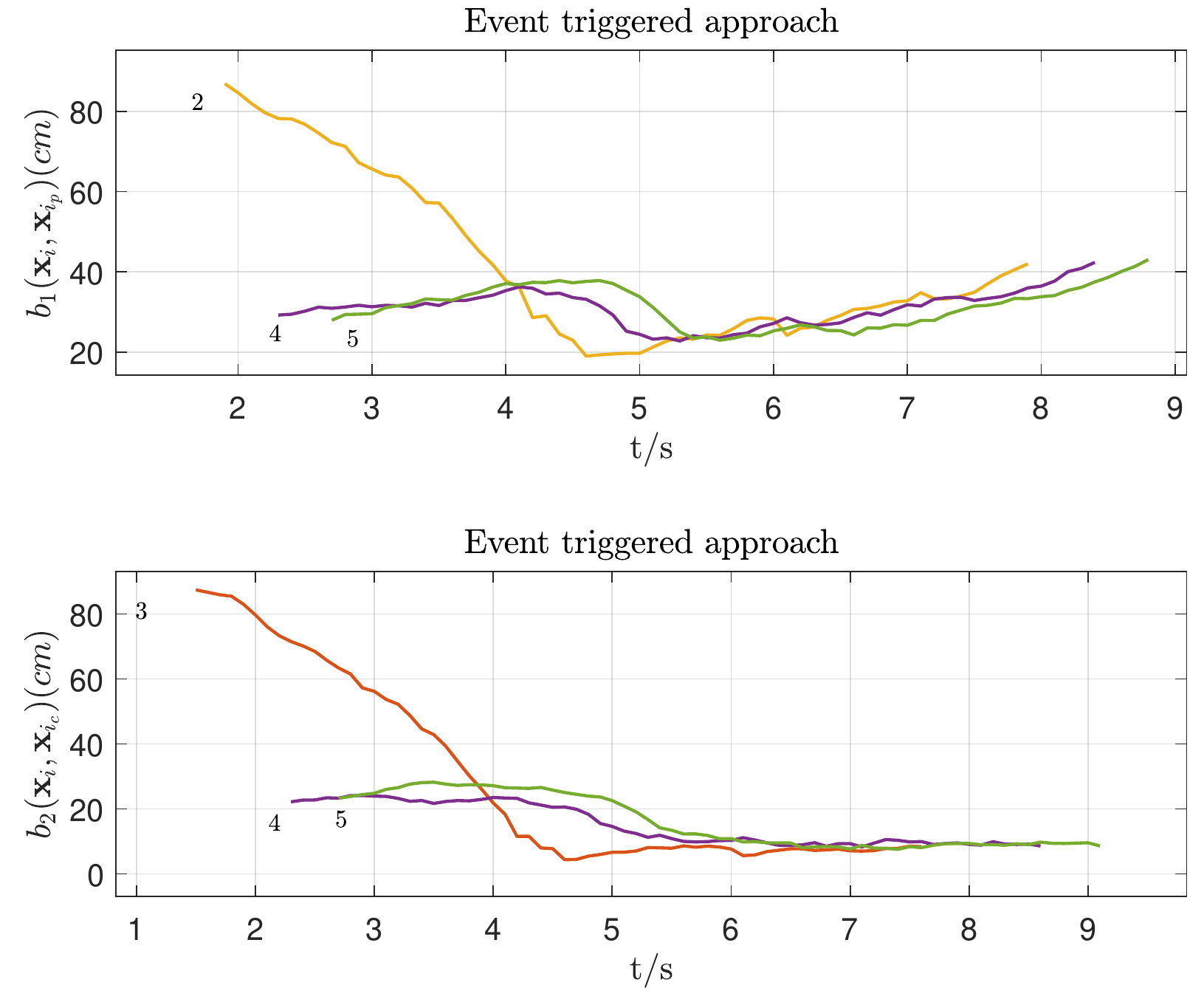}
    \caption{Rear end and safe merging constraints of 5 robots.}
    \label{fig:Rear end and merging 5 CAVs }
\end{figure}

\begin{figure}
    \centering
    \includegraphics[width=0.3\textwidth]{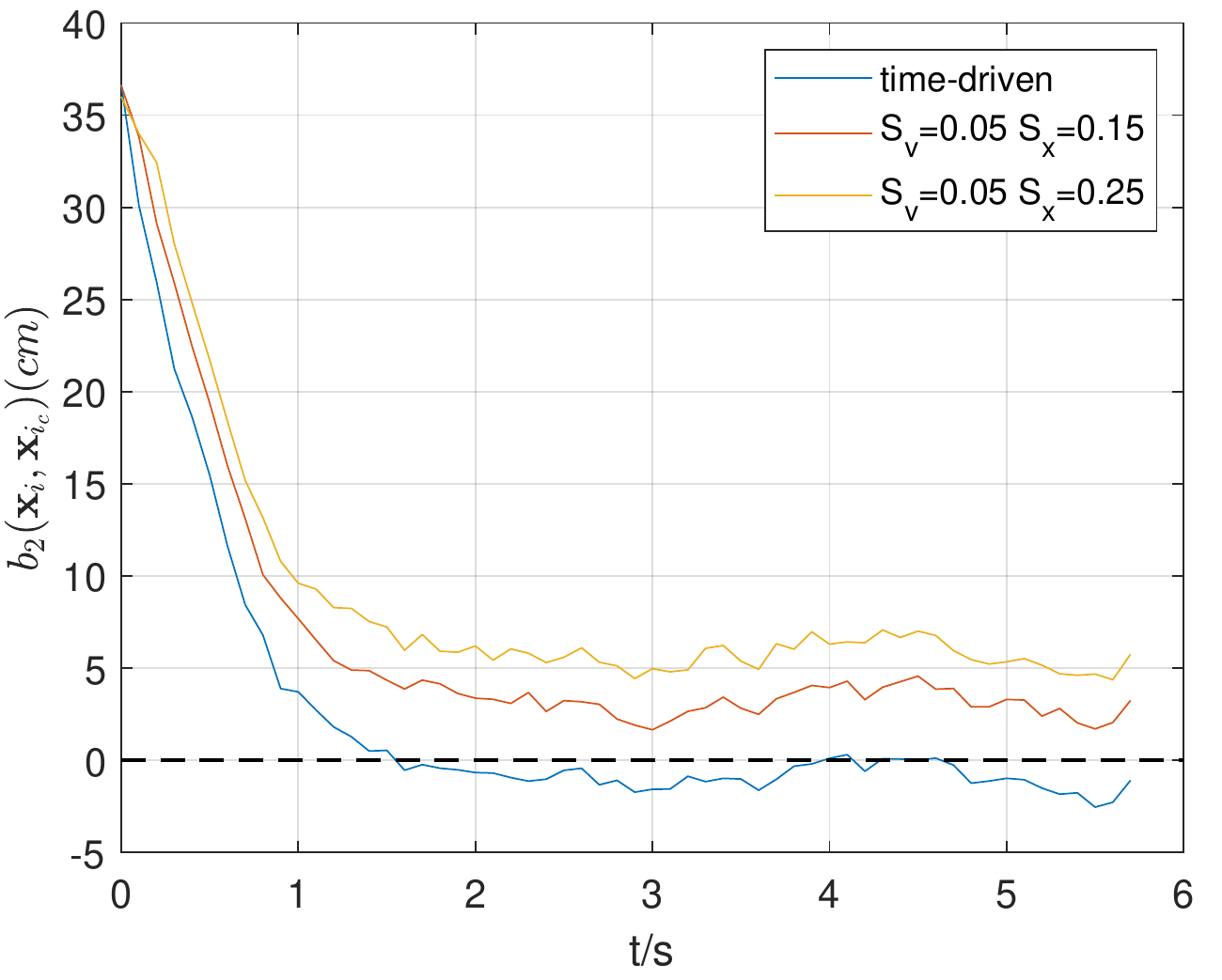}
    \caption{Safe merging constraints of 2 robots under event-triggered scheme with 2 different event bounds and time-driven control schemes. Note the safe merging constraint violation under time-driven control. On the other hand, for both the choice of bounds, safe merging constraint was satisfied at all times by the event-triggered scheme.}
    \label{fig:Merging Constaint} 
\end{figure}

Although simulation results indicate the efficacy of the event-triggered OCBF approach, the results of the lab test bed implementation are significant in two ways: firstly, they demonstrate that the control algorithm can be implemented and executed in real-time, which is essential for safety-critical systems. Secondly, they demonstrate the robustness of the algorithm to the various noise sources present in real-world applications. To illustrate the aforementioned points, two different experiments have been considered. The first experiment involves 5 robots for comprehensive implementation and the results are depicted in Fig. \ref{fig:Rear end and merging 5 CAVs } illustrating the satisfaction of both rear-end safety ($b_1(\textbf{x}_i(t),\textbf{x}_{i_p}(t))$) and merging constraints ($b_2(\textbf{x}_i(t),\textbf{x}_{i_c}(t))$) during the whole experiment duration. In Fig. \ref{fig:Rear end and merging 5 CAVs }, each line specified by the CAV number represents a time plot constraint for that CAV in the CZ. As can be observed, all lines are above zero, therefore, both rear end and merging constraints are satisfied at all times in the CZ. Note that all the CAVs are not present in the plot, as not all the CAVs are constrained. In the second experiment, the initial conditions of the 2 robots (that have to merge together safely coming from two different roads) are deliberately chosen to demonstrate that \emph{the time-driven approach cannot guarantee safety} due to uncertainties and imperfect measurements while the proposed event-triggered scheme with a proper choice of bounds (as discussed in section III) can maintain a safe merging distance ($b_2(\textbf{x}_2(t),\textbf{x}_{1}(t))$) between two robots as depicted in Fig \ref{fig:Merging Constaint}. 

The following web link contains a video of our implementation of the event triggered scheme for a merging roadway in a lab environment: \url{https://www.youtube.com/watch?v=qwhLjEskPS8}. The video contains two scenarios with different arrival sequences where we have used A FIFO passing sequence. In the first scenario, three robots (with indices 1, 2 and 5) arrive from the main road and two robots (with indices 3 and 4) arrive from the side road and they safely merge at the merging point.  For the second scenario, a different arrival order was chosen, whereby three CAVs (with indices 1, 2 and 4) arrive from the main road and two CAVs (with indices 3 and 5) arrive from the side road. The proposed scheme is able to safely merge traffic from both roads as depicted in the video. In this case, CAV 5 has both rear end and merging constraint, to make room for CAV 4 to allow it to merge ahead of it while also staying safe to CAV 3 that is physically preceding it. This represents the most computationally extensive case where a CAV has both constraints thus highlighting the real time feature of the proposed approach. 



\vspace{-2pt}
\section{Conclusion}
The problem of controlling CAVs in conflict areas of a traffic network subject to hard safety constraints by the use of CBFs can be solved through a sequence of QPs. However, the lack of feasibility guarantees for each QP, as well as control update synchronization, motivate an event-triggering scheme. In this paper, we have presented an event-triggered framework for the automated control of merging roadways (which can be generalized to any conflict area) and designed a lab test bed for the cooperative control of CAVs using mobile robots. As part of ongoing work, we will Validate this algorithm in an extensive test bed comprised of multiple conflict points using mobile robots.

\bibliographystyle{IEEETran}

\end{document}